\pdfoutput=1

\documentclass[11pt]{article}

\usepackage[preprint]{acl}

\usepackage{times}
\usepackage{latexsym}

\usepackage[T1]{fontenc}

\usepackage[utf8]{inputenc}

\usepackage{microtype}

\usepackage{inconsolata}

\usepackage{graphicx}
\usepackage{array}
\usepackage{multirow}
\usepackage{amsmath,amssymb}
\usepackage{algorithm}
\usepackage{algpseudocode}
\usepackage{booktabs}
\usepackage{xcolor}

\title{Verbosity-Aware Rationale Reduction: \\
Sentence-Level Rationale Reduction for Efficient and Effective Reasoning
}

\author{
Joonwon Jang\(^{1,5}\) \ 
Jaehee Kim\(^{2}\) \
Wonbin Kweon\(^{3}\) \
Seonghyeon Lee\(^{4}\) \
Hwanjo Yu\(^{1,*}\)\\[1ex] 
POSTECH\(^{1}\), Seoul National University\(^{2}\), University of Illinois Urbana-Champaign\(^{3}\) \\ 
Kyungpook National University\({^4}\), LG AI Research\(^{5}\) \\
\texttt{\{kaoara, hwanjoyu\}@postech.ac.kr}
\texttt{jaehee\_kim@snu.ac.kr}\\
\texttt{wonbin@illinois.edu }
\texttt{sh0416@knu.ac.kr}
}

\begin{document}
\maketitle
\begin{abstract}
Large Language Models (LLMs) rely on generating extensive intermediate reasoning units (e.g., tokens, sentences) to enhance final answer quality across a wide range of complex tasks.
While this approach has proven effective, it inevitably increases substantial inference costs.
Previous methods adopting token-level reduction without clear criteria result in poor performance compared to models trained with complete rationale.
To address this challenge, we propose a novel sentence-level rationale reduction framework leveraging likelihood-based criteria, \textit{verbosity}, to identify and remove redundant reasoning sentences.
Unlike previous approaches, our method leverages \textit{verbosity} to selectively remove redundant reasoning sentences while preserving reasoning capabilities.
Our experimental results across various reasoning tasks demonstrate that our method improves performance by an average of 7.71\% while reducing token generation by 19.87\% compared to model trained with complete reasoning paths.
\end{abstract}

\section{Introduction}
Recent advances in Large Language Models (LLMs) have demonstrated remarkable reasoning capabilities comparable to human cognitive abilities \citep{madaan2024self, shinn2024reflexion, kumar2024traininglanguagemodelsselfcorrect}.
These works demonstrate the capability to solve complex reasoning tasks through explicitly generating extended reasoning paths.
The generation of such paths involves producing explicit reasoning units (e.g., tokens, steps) \citep{yu2024distilling}, which further enhances model performance through iterative prompting \citep{wang2023selfconsistency, yao2023tree}.
Through this iterative generation of explicit reasoning paths, the model refines and expands its thought processes while incorporating strategic planning and continuous cognitive generation \citep{xi2023risepotentiallargelanguage, yang2024buffer}.

While the extensive generation of explicit reasoning units leads to improved performance, it inevitably results in higher inference costs and increased latency \citep{yu2024distilling, wang2024strategicchainofthoughtguidingaccurate}. 
Furthermore, fine-tuning LLMs with complete reasoning paths does not necessarily guarantee enhanced performance \citep{yu2024distilling, deng2024explicitcotimplicitcot, liu2024can}, indicating the necessity for methods that maintain reasoning capabilities while reducing the generation of reasoning units.
Despite this apparent requirement, it remains underexplored how to maintain LLM reasoning capabilities while reducing intermediate reasoning paths across diverse tasks.

Previous methods primarily focused on reducing reasoning paths from two distinct perspectives.
Some studies have proposed training pipelines that leverage augmented datasets, iteratively generated by foundation LLMs, to fine-tune subsequent LLMs \citep{yu2024distilling, liu2024can}.
However, these approaches remain inherently vulnerable due to their significant dependence on the generative capabilities of LLMs.

In response, other works have focused on directly training LLMs without dataset augmentation to reduce explicit reasoning paths.
\citet{deng2023implicitchainthoughtreasoning} introduced a knowledge distillation method to distill explicit reasoning into implicit reasoning through token-level hidden states.
\citet{deng2024explicitcotimplicitcot} adopted tokens as the reasoning unit for reduction and proposed a heuristic method to internalize explicit intermediate rationale tokens while \citet{hao2024training} compressed complete rationales into the predefined number of hidden states of tokens.
However, their methods present a fundamental limitation as they lack sufficient justification for selecting tokens over more linguistically natural units (e.g., sentences) for reduction (Table \ref{tab:reasoning-unit}), and they fail to provide principled criteria for the removal process.
Moreover, their evaluation has primarily focused on synthetic arithmetic reasoning tasks, limiting their applicability to real-world scenarios.

To address these limitations, we propose a novel training method that maintains LLM reasoning performance while systematically reducing redundant reasoning units within the reasoning process.
Our method adopts sentences as fundamental reduction units, establishing more linguistically meaningful boundaries compared to token-level approaches.
Through empirical analysis, we demonstrate that sentences in early rationale steps can introduce redundancy in the LLM's answer generation process.
Inspired by \citet{dong2023statistical}, we introduce the concept of `\textit{verbosity}', a likelihood-based criteria, to identify redundant reasoning sentences.
By incorporating \textit{verbosity} identification into the training process, the model excludes redundant reasoning sentences, thereby reducing intermediate token generation.
Finally, we demonstrate our method's effectiveness and generalizability across various real-world reasoning datasets, showing our method improves performance by an average of 7.71\% while reducing token generation by 19.87\% across various LLMs, and through systematic ablation studies, we analyze the contribution of each proposed component.

\section{Related Works}
\subsection{Performance-Cost Tradeoffs in Reasoning Path Generation}
\label{2.1}

Recent research has demonstrated the critical role of generating iterative and refined reasoning paths in enhancing model reasoning capabilities, albeit at increased computational costs \citep{wang2023selfconsistency, yao2023tree, radha2024iterationthoughtleveraginginner, wang2024strategicchainofthoughtguidingaccurate, madaan2024self, shinn2024reflexion, kumar2024traininglanguagemodelsselfcorrect}.
Self-Consistency \citep{wang2023selfconsistency}, Tree of Thoughts (ToT) \citep{yao2023tree}, and Strategic Chain of Thought (SCoT) \citep{wang2024strategicchainofthoughtguidingaccurate} improve reasoning accuracy through ensemble-based path selection, tree-structured exploration, and adaptive reasoning with an Inner Dialogue Agent, where each approach requires iterative reasoning path generation, resulting in substantial computational overhead.

Concurrently Self-Refine \citep{madaan2024self} and Reflexion-based framework \citep{shinn2024reflexion, kumar2024traininglanguagemodelsselfcorrect} enhance reasoning abilities through iterative feedback-based refinement and reflective path generation, respectively, though both require multiple forward passes through the model.
While the iterative generation and refinement of reasoning paths are essential for achieving optimal performance, they inherently increase inference costs and latency.
Therefore, it is crucial to investigate methods for efficiently generating these paths.

\subsection{Reasoning Path Reduction}

To address the computational costs associated with extensive reasoning paths generation, some lines of work \citep{yu2024distilling, liu2024can} have focused on generating augmented datasets with varying rationale lengths to reduce the generation of reasoning paths.
\citet{yu2024distilling} employs Self-Consistency to generate multiple reasoning paths for dataset augmentation, then fine-tunes the model to produce direct answers.
\citet{liu2024can} developed a heuristic approach to merge reasoning steps and iteratively trained the model to produce shorter reasoning paths, which are then integrated into the progressive training phase.
While these approaches demonstrate empirical effectiveness, they exhibit two fundamental limitations: (1) their substantial dependence on LLM generation capabilities introduces inherent instability, and (2) their objective of reducing reasoning paths necessitates the paradoxical creation of datasets requiring extensive reasoning path generation.

To address these weaknesses, another line of work \citep{deng2023implicitchainthoughtreasoning,deng2024explicitcotimplicitcot, hao2024training} has focused on directly training LLMs without augmented datasets.
Implicit-CoT \citep{deng2023implicitchainthoughtreasoning} implements a multi-model framework where an emulator model is trained to predict the teacher's token-level hidden states, and a student model leverages these predicted states to generate answers.
ICoT-SI \citep{deng2024explicitcotimplicitcot} identifies tokens as reduction units, proposing a method to internalize explicit intermediate rationale tokens by progressively eliminating them from the beginning of the reasoning path within the CoT fine-tuning process.
However, these methods demonstrate limited generalization across diverse datasets as they have been validated exclusively on simple arithmetic reasoning tasks, such as multiplication problems.
This limitation raises concerns about their applicability to real-world scenarios where rationales are expressed in natural language.
Furthermore, they do not explore the adoption of principled criteria and linguistically natural units (e.g., sentences).
Specifically, the token-level reduction approach may eliminate critical information necessary for answer generation or distort the semantic information of the sentence.
Motivated by these limitations in existing reduction approaches, we examine the redundancy of various sentence positions for potential elimination and propose a novel method with principled criteria that can effectively reduce reasoning paths while maintaining their efficacy.

\section{Early Step Rationales are Redundant}
\label{Sec3}

\begin{figure}[ht!]
    \centering
    \scalebox{1}[0.95]{%
        \includegraphics[width=1.0\linewidth]{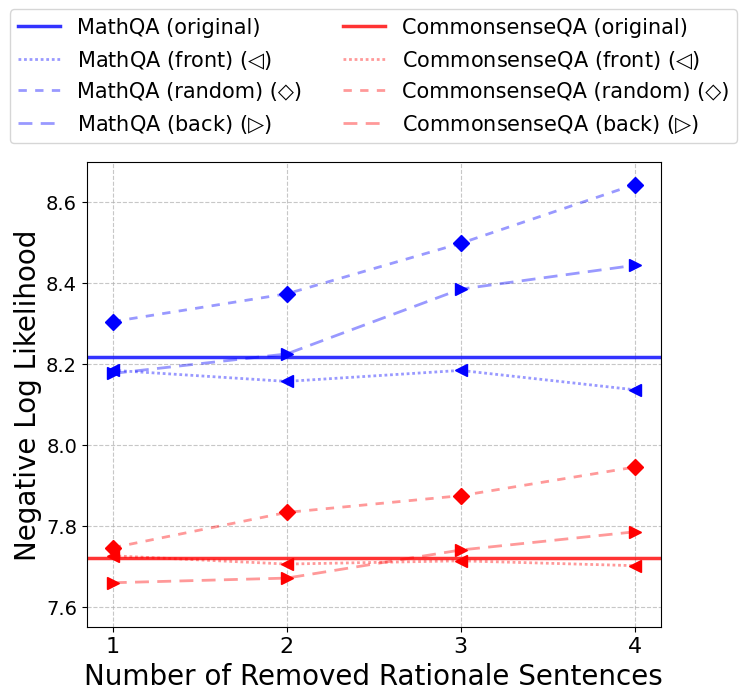}
    }
    \caption{
    NLL differences across varying sizes of $\{r_i\}$. 
    The `original' represents the NLL with the complete rationale, while `front', `random', and `back' indicate that $\{r_i\}$ is sampled from the front, random, and back indices of the full index set, respectively.
    }
    \label{fig:mot-front-remove}
\end{figure}

\subsection{Quanitifying the Redundancy}

Before delving into the method, we first investigate which positions within the rationale sentences should be selected for reduction. 
When the likelihood of the answer remains unchanged after removing a sentence from the full rationales, this indicates that the sentence may be redundant in the reasoning process.
To quantify the redundancy of a sentence, we compute the negative log-likelihood (NLL) for answer $y$ after sentence reduction as follows:

\begingroup
\setlength{\abovedisplayskip}{0pt}
\begin{equation}
\begin{split}
\text{NLL} & = -\log p_\theta(y|R',x), \\
\text{where} \ R' & = R \setminus \{r_i\}_{i \in S}, \ S \subseteq I.
\end{split}
\end{equation}
\endgroup
\noindent Let $R$ denote the complete set of rationale sentences, and $I$ represent the full index set of these sentences.
The subset of indices corresponding to sentences selected for reduction is denoted by $S$, and $R'$ represents the remaining rationale sentences after their removal.
$\{r_i\}_{i \in S}$ denotes the sentences selected for reduction.
For simplicity, we use $\{r_i\}$ without the subset index notation throughout the rest of the paper.

\subsection{Redundancy of Early Reasoning Sentences}

We performed a pilot study to empirically demonstrate the redundancy of leading sentences within the rationales by analyzing the NLL of Mistral 7B \citep{jiang2023mistral7b} across diverse reasoning datasets.
Specifically, we varied the size of $\{r_i\}$ from 1 to 4 and investigated general patterns of sentence removal using a stochastic approach.
To compare different sentence selection configurations, we considered three sampling methods for $\{r_i\}$: \textbf{front}, where initial sentences were prioritized; \textbf{random}, with uniform probabilities; and \textbf{back}, where probabilities progressively increased for later sentences\footnote{For generating $R'$, we assign probabilities $p_k = \frac{N-k+1}{\sum_{i=1}^N i}$ (front), $\frac{1}{N}$ (random), and $\frac{k}{\sum_{i=1}^N i}$ (back) where k=1,...,N denotes sentence position.}. 
Additionally, we computed the NLL for complete rationale sentences (i.e., $-\log p_\theta(y|R,x)$) as a baseline to evaluate the impact of reduction.
As illustrated in Figure \ref{fig:mot-front-remove}, the front ($\triangleleft$) configuration shows only marginal NLL differences relative to complete rationale sentences.
In contrast, removing sentences randomly ($\diamond$) or from the back ($\triangleright$) results in higher NLL as the removed sentences increase, highlighting the importance of the selection of a candidate rationale position strategy for removal in the reasoning and answer prediction process (for additional analysis, see Appendix \ref{appendix:removal}).

\section{Verbosity-Aware Rationale Reduction}
\label{Sec4}

\begin{figure*}[ht!]
    \centering
\includegraphics[width=1.0\linewidth]{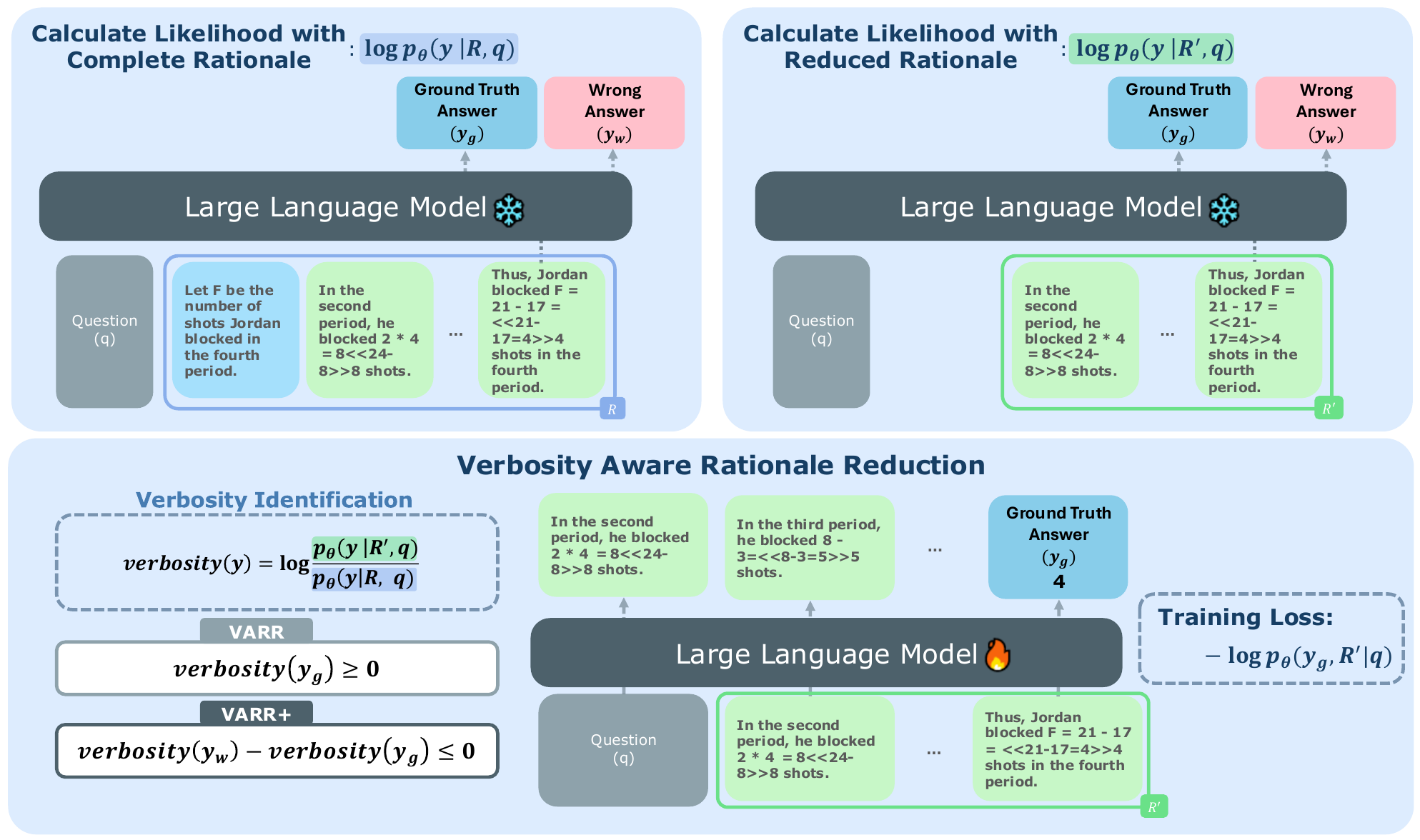}
    \caption{
    Overview of our VARR/VARR+ framework. 
    Initially, we select a candidate sentence from the beginning of the rationale (Section \ref{Sec3}). 
    After selecting the candidate sentence, we evaluate Equations (\ref{eq-6}) and (\ref{eq-12}) by calculating $\textit{verbosity}(y_g)$ and $\textit{verbosity}(y_w)$. 
    If the candidate sentence meets the verbosity evaluation criteria, it is excluded in subsequent training steps. 
    The model then proceeds with training, where the redundant sentence is excluded from the rationale.
    }
    \label{fig:main-method}
\end{figure*}

Based on these observations, we propose the Verbosity-Aware Rationale Reduction (VARR) framework.
In Section \ref{4.1-verbosity}, we introduce the concept of `\textit{verbosity}' as a principled criterion for identifying redundant reasoning sentences. 
Section \ref{4.2-verbosity-identification} elaborates on how we integrate \textit{verbosity} into the reduction process during CoT training.
In Section \ref{4.3-contrast}, we extend the \textit{verbosity} term by incorporating incorrect answers to enhance robustness.
Finally, Section \ref{4.4-CoT-Training} presents the comprehensive VARR framework.

\subsection{Verbosity as Principled Criterion}
\label{4.1-verbosity}
To quantify the redundancy of sentences for potential removal, we introduce the fundamental concept `\textit{verbosity}'. 
Given an input $x$, full rationale $R$, and a reduced rationale $R' = \{r_j\}_{j \in I \setminus \{i\}} $, we quantify the \textit{verbosity} of a sentence $r_i$ on $y$ by computing the difference in Kullback–Leibler divergence (KL-divergence) as follows:

\begingroup
\setlength{\abovedisplayskip}{0pt}
\begin{equation}
\begin{split}
\textit{verbosity}(y) & = D_{KL}(q(y|x) \parallel p_\theta(y|R,x)) \\
& \hspace{1.5em} - D_{KL}(q(y|x) \parallel p_\theta(y|R',x)),
\end{split}
\label{kld}
\end{equation}
\endgroup

\noindent where $q(y|x)$ is the ground truth distribution.
The \textit{verbosity}$(y)$ measures the informational contribution or redundancy of a rationale sentence $r_i$ with respect to answer $y$.
Since $q(y|x)$ is the form of the one-hot vector (i.e., Dirac delta function), we can express the \textit{verbosity}$(y)$ as the log-likelihood ratio between $R$ and $R'$ as follows:

\begingroup
\setlength{\abovedisplayskip}{0pt}
\begin{equation}
\begin{split}
\textit{verbosity}(y_g) & = \left[H_q(p_\theta(y|R,x)) - H(q(y|x))\right] \\
& \hspace{0.7em} - \left[H_q(p_\theta(y|R',x)) - H(q(y|x))\right]\\
& = E_q[-\log p_\theta(y|R,x)] \\
& \hspace{0.7em} + E_q[\log p_\theta(y|R',x)] \\
& = \log \left(\frac{p_\theta(y_g|R',x)}{p_\theta(y_g|R,x)}\right),
\end{split}
\end{equation}
\endgroup

\noindent where $y_g$ denotes the ground truth answer (i.e., $q(y_g|x) = 1$).
$H_q(\cdot)$ and $H(\cdot)$ denote the cross-entropy and the entropy, respectively\footnote{For the sake of explainability, we assume each expression's $y$ is represented by a single token.}.
Intuitvely, a higher value of $\textit{verbosity}(y_g)$ implies that the likelihood of the model generating the ground truth answer increases after removing $r_i$, indicating that its removal is beneficial.

\subsection{Verbosity Identification in CoT Training}
\label{4.2-verbosity-identification}

Given an input sequence, CoT training \citep{nye2021show} aims to train LLMs to generate complete rationale, followed by the ground truth answer:

\begingroup
\setlength{\abovedisplayskip}{0pt}
\begin{equation}
-\log p_\theta(y_g,R|x).
\label{cot-train}
\end{equation}
\endgroup

\noindent During each training step $t$, we evaluate each sentence $r_i$ within $R$ using the following criterion:


\begingroup
\setlength{\abovedisplayskip}{0pt}
\begin{equation}
\textit{verbosity}(y_g)\geq0.
\label{eq-6}
\end{equation}
\endgroup

\noindent Here, we sequentially select $r_i$ starting from the first sentence and construct $R' = \{r_j\}_{j \in I \setminus \{i\}}$, based on our analysis in Section \ref{Sec3}, which indicates that early-stage rationales are more likely to be redundant.
When the verbosity score is non-positive, it indicates that removing $r_i$ from $R$ would impair the model's performance, thus identifying the sentence as essential and preserving it in the rationale.
Detailed training procedures will be described in Section \ref{4.4-CoT-Training}.

\subsection{Contrasting with Wrong Answer}
\label{4.3-contrast}

Inspired by the miscalibrated log-likelihoods between accepted and rejected responses in standalone Supervised Fine-Tuning (SFT) for alignment learning \cite{rafailov2024direct, azar2024general, hong2024orpo}, we examine whether reduced rationales $R'$ lead to inaccurate answer generation by incorporating a wrong answer $y_w$.

Instead of employing the correct answer distribution $q(y|x)$ in Equation (\ref{kld}), we initiate our formula with the wrong answer distribution $q'$ (i.e., $1-q(y|x)$, normalized sum to 1).
Through algebraic manipulation, we derive:

\begingroup
\setlength{\abovedisplayskip}{0pt}
\begin{equation}
\begin{split}
& \hspace{-1em} D_{KL}(q'(y|x) \parallel p_\theta(y|R,x)) \\
& \hspace{1em} - D_{KL}(q'(y|x) \parallel p_\theta(y|R',x)) \\
& \hspace{-1em} = \left[H_w(p_\theta(y|R,x)) - H(q'(y|x))\right] \\
& \hspace{1em} - \left[H_w(p_\theta(y|R',x)) - H(q'(y|x))\right],
\end{split}
\end{equation}
\endgroup

\noindent where $H_w(\cdot)$ denotes the cross-entropy calculated with the wrong answer distribution $q'$.
Due to the impracticality of computing the expectation over the entire space of $V-1$ wrong answers (where $V$ is the vocabulary size), we sample $K$ incorrect answers for the following estimations:

\begingroup
\setlength{\abovedisplayskip}{0pt}
\begin{equation}
\begin{split}
& E_w[-\log p_\theta(y|R,x)] + E_w[\log p_\theta(y|R',x)] \\
& = {E}_{w}\Big[\log\frac{p_\theta(y|R',x)}{p_\theta(y|R,x)}\Big] \\
& \approx \frac{1}{K}\sum_{k=1}^K \log\frac{p_\theta(y_w^{(k)}|R',x)}{p_\theta(y_w^{(k)}|R,x)},
\end{split}
\end{equation}
\endgroup

\noindent where $\{ y_w^{(k)} \}_{k \in \left[K\right]}$ is sampled from the in-batch negatives depending on the dataset.
Consequently, $\textit{verbosity}(y_w)$ is computed as:

\begingroup
\setlength{\abovedisplayskip}{0pt}
\begin{equation}
    \textit{verbosity}(y_w) = \frac{1}{K}\sum_{k=1}^K \log\frac{p_\theta(y_w^{(k)}|R',x)}{p_\theta(y_w^{(k)}|R,x)}.    
\end{equation}
\endgroup

\noindent Since computational constraints necessitate sampling incorrect answers to calculate $\textit{verbosity}(y_w)$, 
we evaluate the effectiveness of removal by comparing $\textit{verbosity}(y_w)$ against $\textit{verbosity}(y_g)$ rather than solely using $\textit{verbosity}(y_w)$ as follows:

\begingroup
\setlength{\abovedisplayskip}{0pt}
\begin{equation}
    \textit{verbosity}(y_w) - \textit{verbosity}(y_g) \leq 0.
    \label{eq-12}
\end{equation}
\endgroup
 \noindent When both conditions $\textit{verbosity}(y_g) \geq 0$ and $\textit{verbosity}(y_w) - \textit{verbosity}(y_g) \leq 0$ are satisfied, it indicates that removing $r_i$ from $R$ not only improves the model's performance but also increases its preference for the ground truth answer over incorrect answers, supporting the removal of $r_i$.

\subsection{CoT Training with Rationale Reduction}
\label{4.4-CoT-Training}

In our framework, the model is trained using Equation (\ref{cot-train}) for predefined warm-up stage to inject its reasoning capabilities.
Subsequently, at each training step $t$, we evaluate each sentence $r_i$ sequentially from the first sentence in $R$, using either Equation (\ref{eq-6}) alone (denoted as VARR) or the combination of Equations (\ref{eq-6}) and (\ref{eq-12}) (denoted as VARR+).
Sentences that satisfy these respective criteria are identified for removal and excluded from subsequent training steps.
The maximum removable number of sentences at each training step $t$ is determined based on a linear schedule, adopting ICoT-SI \citep{deng2024explicitcotimplicitcot}'s setting as follows:

\begingroup
\setlength{\abovedisplayskip}{0pt}
\begin{equation}
r(t) = \lfloor N_t \cdot (t / T) \rfloor,
\end{equation}
\endgroup

\noindent where $T$ represents the total number of training steps, $N_t$ is the total number of rationale sentences at step $t$, and $r(t)$ indicates the maximum number of sentences that can be removed at that step.
Note that unlike ICoT-SI, which enforcely removes a predefined number of tokens during training, our method preserves the essential reasoning steps by employing principled removal criteria.




\section{Experiments}
\subsection{Training Configuration}

\begin{figure*}[ht!]
    \centering
    \includegraphics[width=1.0\linewidth, height=0.4\textheight]{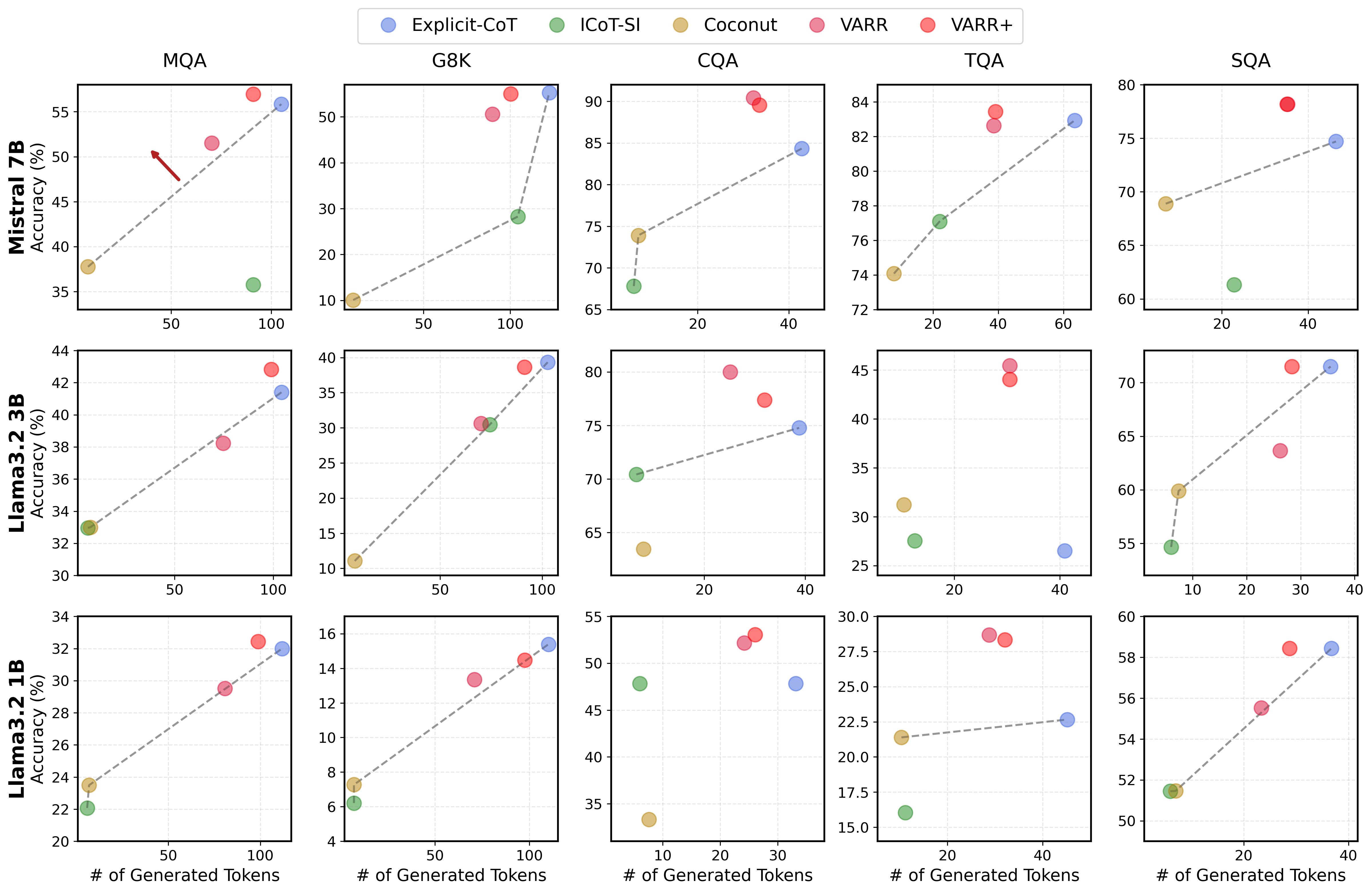}
    \caption{
    Pareto plot of accuracy versus the number of generated tokens. 
    The gray dotted lines connect the Pareto frontiers of the baselines, and our \textbf{VARR (or VARR+) consistently outperforms the pareto frontiers} across all subplots. 
    While ICoT-SI and Coconut substantially trade-off accuracy for efficiency, VARR/VARR+ maintains high accuracy while reducing generated tokens, demonstrating its superior efficiency-performance balance.
    }
    \label{fig:main-results}
\end{figure*}

\paragraph{Datasets}

We conducted experiments across two categories to provide a comprehensive evaluation of VARR’s versatility and effectiveness, in contrast to prior research that predominantly focuses on simple arithmetic tasks like multi-digit multiplication \citep{deng2023implicitchainthoughtreasoning,deng2024explicitcotimplicitcot}.
Initially, we evaluate with arithmetic reasoning tasks, including datasets like MathQA (MQA; \citealt{amini2019mathqa}) and GSM8K (G8K; \citealt{cobbe2021trainingverifierssolvemath}). 
We also examine the performance of our method on commonsense reasoning tasks, employing datasets including CommonsenseQA (CQA; \citealt{talmor2019commonsenseqa}), TriviaQA (TQA; \citealt{joshi2017triviaqa}), and StrategyQA (SQA; \citealt{geva2021did}).
Note that unlike previous works \citep{deng2023implicitchainthoughtreasoning, deng2024explicitcotimplicitcot}, we do not synthesize training data (especially the intermediate steps) to validate the generalizability and applicability of our method.

\paragraph{Models}
 We trained Mistral 7B \citep{jiang2023mistral7b} as our base model for comparisons. 
We also trained a series of Llama3.2 models \citep{grattafiori2024llama3herdmodels} scaling from 1B to 3B to demonstrate our method's generalization capabilities.

\paragraph{Implementation Details}

The warm-up stage is set to 0.1 of the total training steps, and its impact is analyzed in Section \ref{ab-warm-up}. 
In addition, an optimizer is reinitialized at the beginning of each epoch to stabilize the model training inspired by \citet{deng2024explicitcotimplicitcot}'s setting, with its effects described in Appendix \ref{appendix:optimizer}. 
All methods are trained for 5 epochs for fair comparison, with detailed implementation of VARR/VARR+ provided in Appendix \ref{appendix:varr-imp}.

\subsection{Baselines}
We compared our method against the following baselines: \textbf{Explicit-CoT} \citep{nye2021show}, where the model is finetuned with explicit chain-of-thought reasoning; \textbf{ICoT-SI} \citep{deng2024explicitcotimplicitcot}, where the model is fine-tuned using a linear token elimination schedule; and \textbf{Coconut} \citep{hao2024training}, where the model is fined-tuned to compress rationales into a predefined number of token hidden states.
We excluded Implicit-CoT \cite{deng2023implicitchainthoughtreasoning} from our evaluation due to its substantial computational demands, specifically requiring three models to be trained simultaneously on a single GPU.
Moreover, \citet{deng2024explicitcotimplicitcot} demonstrated that this method achieves a lower performance compared to ICoT-SI. 
Given that both our baselines and VARR aim to \textit{maintain Explicit-CoT's performance while reducing the number of generated tokens}, we establish Explicit-CoT's performance as our primary baseline for comparison. 
All baselines were trained on a single A100-80GB GPU, and detailed training configurations for each method are provided in Appendix \ref{appendix:exp-details}.

\subsection{Evaluation}

We employ two evaluation metrics: First, we evaluate the accuracy of each method in generating the final answer for the respective tasks. 
Second, we count the generated tokens to evaluate reasoning efficiency while maintaining performance.

\subsection{Main results}

In Figure \ref{fig:main-results}, we present the results for each reasoning task across different models.
VARR/VARR+ achieves comparable or superior performance compared to Explicit-Cot across most datasets while reducing the average token generation.
Specifically, VARR+ significantly increases performance by an average of 7.71\% across all datasets and models, while improving efficiency by reducing token generation by 19.37\% on average.


It is noteworthy that these findings contrast with ICoT-SI, demonstrating that performance can be improved while reducing the number of generated tokens.
This suggests that existing reasoning data contains unnecessary reasoning sentences that may harm performance.
Furthermore, effective reasoning can be achieved by selectively removing sentences based on appropriate criteria.

However, the baselines exhibit performance degradation compared to Explicit-CoT. 
For ICoT-SI, we observe an average performance decline of 21.98\%, while Coconut shows a degradation of 25.20\%, demonstrating their imbalanced trade-off in efficiency.
These results suggest that heuristic reasoning reduction approaches do not effectively induce implicit reasoning within the model as discussed by ICoT-SI and Coconut.
This indicates that identifying and retaining appropriate reasoning units through principled criteria is crucial for maintaining performance in practical applications.
A more detailed discussion of the impact of the choice of reduction unit and criterion is provided in Section \ref{ab-unit}.

Furthermore, qualitative analysis (refer to Appendix \ref{appendix:qualitative}) confirms that ICoT-SI and Coconut fail to generate valid reasoning paths for answer generation, while VARR+ produces concise, yet effective reasoning paths that lead to correct answers. 
Additionally, the incorporation of incorrect answers in VARR+ resulted in performance improvements across most datasets. 
While VARR alone enhanced generation efficiency, VARR+ effectively preserves rationales that help calibrate the probability distribution between correct and incorrect answers, contributing to improved training robustness and stability.



\subsection{Ablation Studies}

In this section, we conduct ablation studies to empirically validate our method. 
All experiments are implemented using Mistral 7B due to its higher base capacity compared to other models.

\subsubsection{Identifying Appropriate Units for Removal}
\label{ab-unit}

\begin{table}[ht!]
\centering
\renewcommand{\arraystretch}{1.0}
\resizebox{\columnwidth}{!}{
\begin{tabular}{cccccc}
\toprule
 & MQA & G8K & CQA & TQA & SQA \\
\midrule
\multirow{2}{*}{Exp-CoT} & 55.84 & \textbf{55.26} & 84.33 & 82.94 & 74.70 \\ [-2.7pt]
                              & (105.02) & (122.54) & (42.84) & (63.46) & (46.47) \\
\midrule
\multirow{2}{*}{ICoT-SI} & 35.84 & 28.27 & 67.82 & 77.09 & 61.33 \\ [-2.7pt]
                              & (113.55) & (104.41) & (6.0) & (22.03) & (22.86) \\
\midrule
\multirow{2}{*}{VARR-Tok}        & 46.79 & 47.53 & 82.60 & 67.14 & 71.22 \\ [-2.7pt]
                              & (90.39) & (94.16) & (25.07) & (38.10) & (31.92) \\
\midrule
\multirow{2}{*}{VARR-Sent}     & \textbf{56.95} & 54.98 & \textbf{89.56} & \textbf{83.45} & \textbf{78.19} \\ [-2.7pt]
                              & (91.04) & (100.38) & (33.55) & (39.17) & (35.12) \\
\bottomrule
\end{tabular}
}
\caption{
Analysis across various reasoning reduction units and the application of principled criteria.
Each row presents accuracy in the first line, with average generated tokens shown in parentheses in the second line.
}
\label{tab:reasoning-unit}
\end{table}

In this section, we empirically investigate the necessity of our criteria and demonstrate why sentences are more effective than tokens as reasoning reduction units.
While ICoT-SI removes tokens without specific criteria, we first apply VARR+ at the token level (denoted as VARR-Tok) to assess whether tokens can serve as effective reduction units when combined with our criteria.
As shown in Table \ref{tab:reasoning-unit}, VARR+ applied at the token level achieves an average performance gain of 24.74\% compared to ICoT-SI, demonstrating that our principled criteria contribute to robust performance.
Furthermore, expanding the reduction units from tokens to sentences (denoted as VARR-Sent) yields an additional performance gain of 15.98\% over VARR-Tok.
These findings highlight that sentences provide natural and effective boundaries for the reduction process.

\subsubsection{Analyzing the Impact of Sentence Position on Removal Efficacy}

\begin{table}[ht!]
\centering
\renewcommand{\arraystretch}{1.0}
\resizebox{\columnwidth}{!}{
\begin{tabular}{cccccc}
\toprule
 & MQA & G8K & CQA & TQA & SQA \\
\midrule
\multirow{2}{*}{Exp-CoT} & 55.84 & \textbf{55.26} & 84.33 & 82.94 & 74.70 \\ [-2.7pt]
                              & (105.02) & (122.54) & (42.84) & (63.46) & (46.47) \\
\midrule
\multirow{2}{*}{No Rule} & 35.06 & 25.85 & 74.78 & 74.14 & 72.38 \\[-2.7pt]
 & (59.92) & (63.56) & (23.56) & (6.84) & (23.38) \\
\midrule
\multirow{2}{*}{Random} & 55.34 & 52.31 & 83.47 & 79.62 & 73.83 \\[-2.7pt]
 & (99.13) & (115.59) & (49.19) & (44.98) & (50.51) \\
\midrule
\multirow{2}{*}{Back} & 49.71 & 48.52 & 85.21 & 70.0 & 74.41 \\[-2.7pt]
 & (92.64) & (103.25) & (19.86) & (36.22) & (18.17) \\
\midrule
\multirow{2}{*}{Front} & \textbf{56.95} & 54.98 & \textbf{89.56} & \textbf{83.45} & \textbf{78.19} \\[-2.7pt]
 & (91.04) & (100.38) & (33.55) & (39.17) & (35.12) \\
\bottomrule
\end{tabular}}
\caption{
Performance across 5 different reasoning tasks, evaluated with different sentence position removal. 
Each row presents accuracy in the first line, with average generated tokens shown in parentheses in the second line.
}
\label{tab:reasoning-position}
\end{table}

In Section \ref{Sec3}, we demonstrated that gradually removing sentences from random and back positions can degrade model performance. 
To further explore this finding and assess the robustness of removing sentences from the front position, we conducted experiments with unguided random sentence removal (denoted as No Rule) and applied VARR+ with random position and reverse sentence order (denoted as Random and Back, respectively). 
As shown in Table \ref{tab:reasoning-position}, unguided random sentence removal resulted in a 25.30\% decrease in performance relative to our method, highlighting the critical role of the verbosity evaluation even after selecting sentences as units of reduction. 
Furthermore, Random and Back strategies exhibited an average 7.50\% performance degradation relative to our method.
These results further support our observation that earlier sentences in the reasoning path tend to contain more redundancy, and their prioritized removal effectively balances rationale reduction while maintaining reasoning performance.

\begin{table}[ht!]
\centering
\renewcommand{\arraystretch}{1.0}
\resizebox{\columnwidth}{!}{
\begin{tabular}{cccc}
\toprule
 & G8K & SQA & TQA \\
\midrule
\multirow{2}{*}{Exp-CoT} & \textbf{55.84} & 74.70 & 82.94 \\ [-2.7pt]
                              & (122.54) & (46.47) & (63.46) \\
\midrule
\multirow{2}{*}{RAND-SENT REMOVAL} & 25.85 & 72.38 & 74.14 \\[-2.7pt]
 & (63.56) & (23.38) & (6.84) \\
\midrule
\multirow{2}{*}{FRONT-SENT REMOVAL (1)} & 38.36 & 72.96 & 79.74 \\[-2.7pt]
 & (87.43) & (24.34) & (22.73) \\
\midrule
\multirow{2}{*}{FRONT-SENT REMOVAL (2)} & 23.50 & 73.83 & 76.79 \\[-2.7pt]
 & (47.64) & (9.97) & (8.32) \\
\midrule
\multirow{2}{*}{VARR+} & 54.98 & \textbf{78.19} & \textbf{83.45} \\[-2.7pt]
 & (100.38) & (35.12) & (39.17) \\
\bottomrule
\end{tabular}}
\caption{
Performance across different front sentence removal strategies.
RAND-SENT-REMOVAL refers to the Random strategy, while $(\cdot)$ in the FRONT-SENT REMOVAL indicates the number of enforcedly removed front sentences.
Each row presents accuracy in the first line, with average generated tokens shown in parentheses in the second line.}
\label{tab:front-sent-enf}
\end{table}

To further validate whether VARR enhances robustness compared to enforced removal of front sentences, we conducted additional experiments where the first and second sentences were enforced to be removed during the initial two training epochs. 
As shown in Table \ref{tab:front-sent-enf}, enforced removal of front sentences leads to substantial performance degradation, while VARR demonstrates its contribution to preserving robustness.

\subsubsection{Varying the Warm-up Ratio}
\label{ab-warm-up}

\begin{figure}[ht!]
    \centering
    \renewcommand{\arraystretch}{0.9}
    \includegraphics[width=0.48\textwidth,height=0.5\textheight,keepaspectratio]{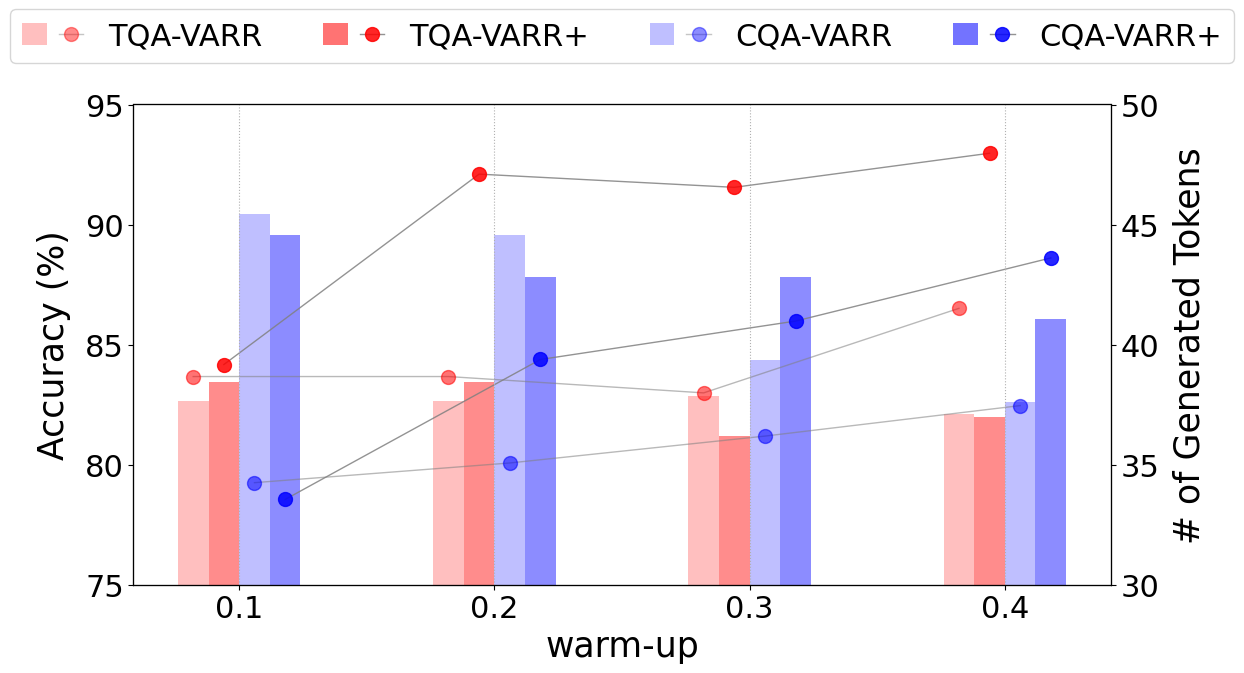}
    \caption{
    Accuracy (barplot) and the average generated token (marker) across various warm-up stages on TriviaQA and CommomsenseQA.
    }
    \label{fig:warm_up}
\end{figure}

We evaluated our method against various warm-up stages on the TriviaQA and CommonsenseQA datasets. 
As the duration of the warm-up stages increases, the model becomes more fitted to the non-reduction dataset, which inhibits VARR's ability to eliminate redundant sentences from the reasoning path.
Consequently, as illustrated in Figure \ref{fig:warm_up}, longer warm-up periods result in an increase in generated tokens and a decrease in accuracy.
These results suggest that 0.1 training steps provide sufficient time to inject reasoning abilities while enabling the systematic reduction of redundant reasoning sentences during the learning process.
The zero warm-up configuration demonstrates comparable performance, though 0.1 warm-up steps are slightly better (detailed in Appendix \ref{appendix:zero-warm}).
Therefore, we set 0.1 as the default setting for the warm-up steps.

\subsubsection{Removal Ratio Analysis}
\label{ab-removal-analysis}

\begin{figure}[ht!]
    \centering
    \renewcommand{\arraystretch}{0.8}
    \includegraphics[width=0.48\textwidth,height=0.5\textheight,keepaspectratio]{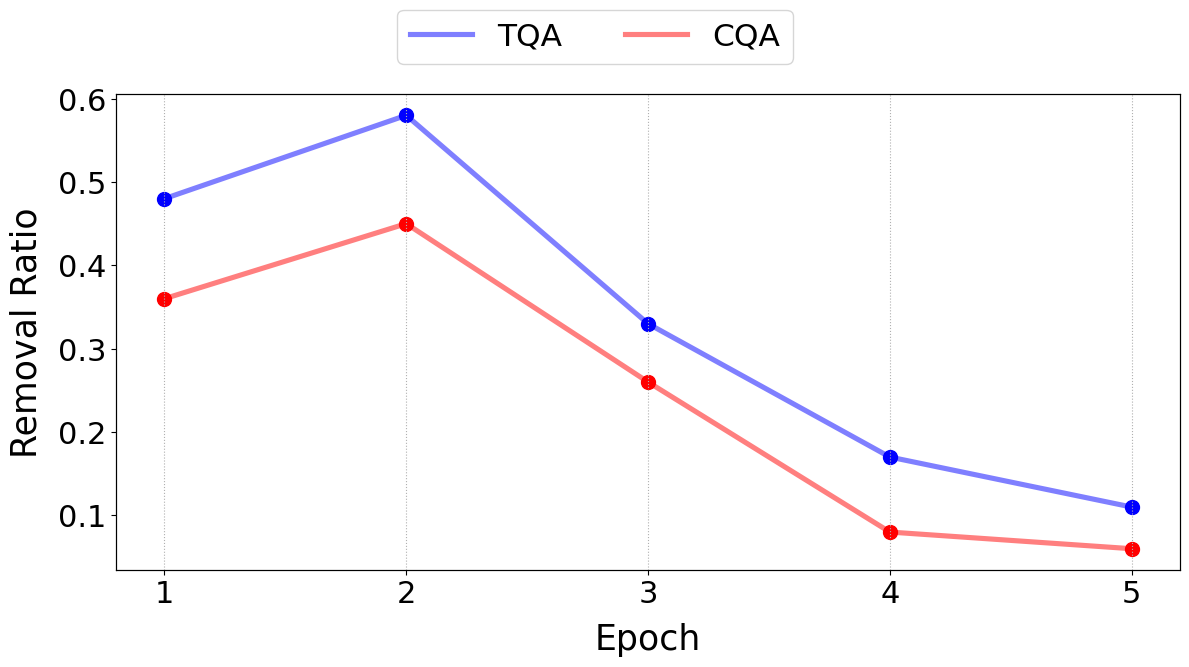}
    \caption{
    Removal ratio of redundant sentences during training.
    The y-axis shows the removal ratio, calculated as the number of removed sentences divided by the maximum potential removal sentences (${size(\{r_i\})/r(t)}$).
    The x-axis represents training epochs.
    }
    \label{fig:removal}
\end{figure}

In Figure \ref{fig:removal}, we analyze the actual amount of rationale sentences removed during training.
We examine it by calculating the removal ratio, the proportion of actual removed sentences to the maximum potential removal sentences $r(t)$. 
Our analysis indicates that not all sentences designated for maximum removal range are always eliminated during the training process. 
Notably, a significant proportion of redundant sentences are removed in the early stages of training, with fewer sentences being removed as the model progresses through the middle to later stages, thereby stabilizing its training. 
A similar trend is observed across other datasets, as detailed in Appendix \ref{appendix:removal-ratio}.

\section{Conclusion}
In this work, we propose the sentence-level rationale reduction framework VARR and empirically demonstrate that models trained with non-redundant rationales achieve enhanced efficiency.
We address the lack of principled criteria for identifying redundant sentences during training by developing a reduction framework that not only preserves the model's reasoning capabilities but also reduces the likelihood of generating incorrect answers.
Our experiments show that VARR can efficiently handle a diverse range of tasks with fewer generated tokens, without sacrificing its accuracy. 
This work contributes novel insights to rationale reduction research, contributing to the efficient reasoning elicitation in language models.

\newpage

\section*{Limitations}
While our work provides novel insights into rationale reduction research, our experiments were primarily conducted using a relatively small large language model and limited batch size, constrained by computational costs (i.e., a single A100-80GB GPU).
Additionally, for the same reasons, it was not feasible to test the model with datasets featuring long sequences in both queries and rationales \citep{reddy2024docfinqa, yu2024metamath}. 
Nevertheless, given the systematic design principles underlying the VARR/VARR+ frameworks, we believe their effectiveness would extend to larger-scale implementations.
Furthermore, we reserve the application of VARR/VARR+ in iterative reasoning path generation and refinement/reflexion-based evaluation discussed in Section \ref{2.1} for future work.


\section*{Ethical Considerations}
Our work explores how LLMs can maintain their reasoning performance while improving efficiency.
To this end, we conducted verbosity-aware rationale reduction (i.e., reasoning sentence pruning)-based CoT fine-tuning, requiring computational resources comparable to standard CoT fine-tuning.
Additionally, we used only open-source LLMs and publicly available reasoning datasets with minimal preprocessing using \texttt{gpt4o-mini}'s \citep{gpt4o-mini} API.
Therefore, we do not anticipate significant ethical issues arising from our work.
On the contrary, we believe future works could leverage our analysis to reduce computational overhead in CoT inference settings.

\section*{Acknowledgements}
We sincerely appreciate to Keonwoo Kim and Hyowon Cho for their sharp feedback during writing, Sangyeop Kim, Yukyung Lee, and Minjin Jeon for their continuous encouragement and inspiration during our study sessions, and Eunbi Choi for her invaluable help during the rebuttal process.
This work was supported by Institute of Information \& communications Technology Planning \& Evaluation (IITP) grant funded by the Korea government(MSIT) (No.RS-2019-II191906, Artificial Intelligence Graduate School Program(POSTECH)), Institute of Information \& communications Technology Planning \& Evaluation (IITP) grant funded by the Korea government (MSIT) (No.2018-0-00584, (SW starlab) Development of Decision Support System Software based on Next-Generation Machine Learning), National Research Foundation of Korea (NRF) grant funded by the Korea government (MSIT) (No. RS-2024-00335873), and National Research Foundation of Korea (NRF) grant funded by the Korea government (MSIT) (No. RS-2023-00217286).

\bibliography{custom}

\appendix

\section{Additional Datasets Analysis}
\label{appendix:removal}

\begin{figure}[h!]
    \centering
    \includegraphics[width=1.0\linewidth]{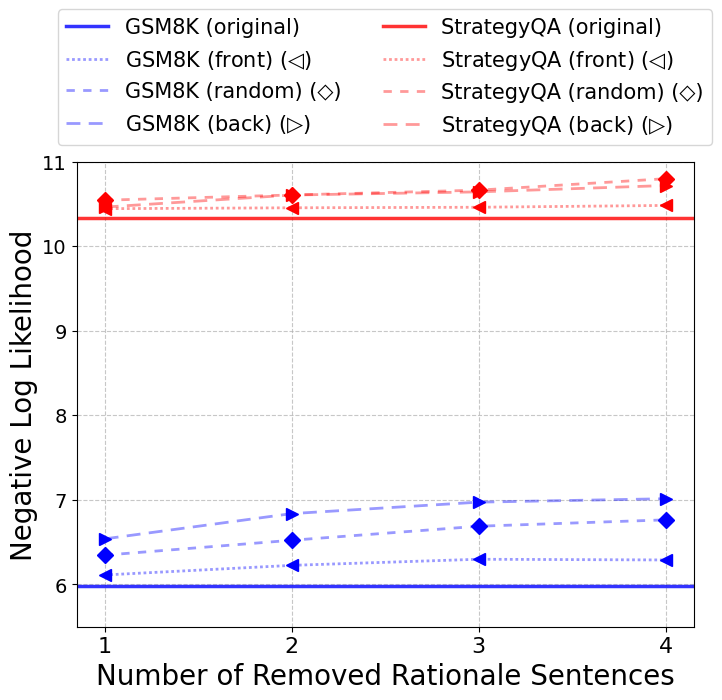}
    \caption{NLL differences across varying sizes of $\{r_i\}$. 
    The `original' represents the NLL for the full rationale, while `front', `random', and `back' indicate that ${r_i}$ are sampled from the front, random, and back indices of the full index set, respectively.}
    \label{fig:app-motivation}
\end{figure}

As shown in Section \ref{Sec3}, the NLL tends to rise with increasing size of $\{r_i\}$ in random and back configurations, compared to the front configuration, as illustrated in Figure \ref{fig:app-motivation}, this trend is consistent across both the GSM8K and StrategyQA datasets.

\section{VARR Implementation}
\label{appendix:varr-imp}

The detailed implementation of VARR+ is outlined in Algorithm \ref{app-imp}.
After the warm-up stages, each datum in the current batch is evaluated using the verbosity equation—specifically, using only Equation~\ref{eq-6} in VARR and both Equations~\ref{eq-6} and~\ref{eq-12} in VARR+. 
As mentioned in Section \ref{Sec3} and \ref{Sec4}, the process assesses the redundancy of each sentence in data starting from the first index during every single epoch.
Therefore, all data can potentially shorten the length of the rationale progressively.

\begin{algorithm}[ht!]
\small
\caption{Training Procedure of VARR+}
\begin{algorithmic}[1]
    \State $\mathcal{D}$: Training dataset
    \State $\mathcal{B}$: Training batch
    \State $E$: Total training epochs
    \State $S$: Total training steps per epoch
    \State $T = E \times S$: Total training steps
    \State $T_{warmup} = 0.1 \times T$: Warm-up steps
    \State $R_B$: A buffer to store removed sentences
    \State $r(t)$: Maximum number of removable sentences at $t$
    \State $N_t$: Number of rationale sentences at $t$
    \State $\theta$: Trained model parameters

    \For{epoch = 1 to E}
        \For{step = 1 to S}
            \State Sample training batch $\mathcal{B}$ from $\mathcal{D}$
            \State $t = (epoch - 1) \times S + step$ 

            \If{$t \leq T_{warmup}$}
                \State pass 
            \EndIf

            \For{each $d \in \mathcal{B}$}
                \State $R_B \gets \{\}$
                \For{i = 1 to $N_t$}
                    \If{$r_i$ satisfies Equations~\ref{eq-6} and~\ref{eq-12}}
                        \State remove $r_i$ from $d$
                        \State Add $r_i$ to $R_B$
                        \If{$|R_B| \geq r(t)$}
                            \State \textbf{break}
                        \EndIf
                    \EndIf
                \EndFor
            \EndFor

            \State Forward pass
            \State Backward pass and update $\theta$
        \EndFor
        \State Reinitialize optimizer
    \EndFor
\end{algorithmic}
\label{app-imp}
\end{algorithm}

\section{Additional Experimental Details}
\label{appendix:exp-details}

For all experiments, we employ the AdamW optimizer \cite{loshchilov2018decoupled}, configured with a weight decay of 0.005.
For the Mistral 7B model, we utilize an effective batch size of 12 with gradient accumulation set to 3, while the smaller Llama3.2 models use an effective batch size of 15. 
For Coconut \citep{hao2024training}, we set the max\_latent\_stage to 5 while maintaining all other hyperparameters as their default repository values unless otherwise mentioned.
A constant learning rate of $5 \times 10^{-6}$ is applied across all datasets, with bfloat16 precision.
For Multiple-Choice and True/False tasks, complete sets of non-correct labels are employed to configure in-batch negatives to enhance the stability of the verbosity evaluation. 
To ensure a fair comparison, all baselines and methods are trained on a single A100 GPU with 80GB of memory for up to 5 epochs or 36 hours, whichever is reached first and experimented with single run evaluation (note that our setting is different from prompting/inference only setting).
Regarding licensing, Mistral 7B is licensed under Apache License, Version 2.0, while Llama3.2 is governed by the \href{https://github.com/meta-llama/llama-models/blob/main/models/llama3_2/LICENSE}{Llama 3.2 Community License}.

\section{Details and Statistics of Datasets}
\label{appendix:data-details}

\begin{table*}[ht!]
\renewcommand{\arraystretch}{1.0}  
\begin{center}
\resizebox{\linewidth}{!}{  
\begin{tabular}{l|c|c|c|c|c|c}
\hline
\textbf{Dataset} & \textbf{Reasoning Task} & \textbf{Source} & \textbf{Answer Format} & \textbf{\# TRAIN.} & \textbf{\# VALID.} & \textbf{\# TEST.} \\
\hline
GSM8K \citep{cobbe2021trainingverifierssolvemath} & Arithmetic & \citet{cobbe2021gsm8k} & Number & 7000 & 473 & 1319 \\
MathQA \citep{amini2019mathqa} & Arithmetic & \citet{amini-etal-2019-mathqa} & Multi-choice & 29837 & 4475 & 2985 \\
TriviaQA \citep{joshi2017triviaqa} & Commonsense & \citet{kim2023cot} & Natural Language & 8844 & 552 & 1659 \\
CommonsenseQA \citep{talmor2019commonsenseqa} & Multi-choice &  \citet{kim2023cot} & Multi-choice & 609 & 38 & 115 \\
StrategyQA \citep{geva2021did} & Commonsense & \citet{sileo-2024-tasksource-large} & T/F & 1832 & 114 & 344 \\
\hline
\end{tabular}
}
\end{center}
\vspace{-0.3cm}  
\caption{
Comprehensive statistics of the datasets used in our experiments are provided. 
GSM8K and MathQA are sourced directly from their original datasets, while the remaining datasets were obtained from \citet{kim2023cot} and \citet{sileo-2024-tasksource-large} to access complete reasoning paths. GSM8K, StrategyQA, and CommonsenseQA are licensed under the MIT License, whereas MathQA and TriviaQA are distributed under the Apache License, Version 2.0.
}
\label{app-tab:datasets}
\end{table*}

For our experimental analysis, we carefully selected a diverse set of five datasets used in prior works \citep{deng2023implicitchainthoughtreasoning, deng2024rephraserespondletlarge, liu2024can, yu2024distilling, yin2024reasoning}. 
To ensure explicit sentence boundaries, all datasets were preprocessed using \texttt{gpt-4o-mini} \citep{gpt4o-mini} to establish clear sentence demarcation (e.g., \textcolor{red!50}{`He bikes 20*2=<<20*2=40>>40 miles each day for work So he bikes 40*5=<<40*5=200>>200 miles for work'} becomes \textcolor{red!80}{`He bikes 202=<<202=40>>40 miles each day for work. So he bikes 405=<<405=200>>200 miles for work'}). 
Table \ref{app-tab:datasets} comprehensively outlines each dataset, including its source and the size of the training, validation, and test samples.

\begin{table}[h!]
\centering
\renewcommand{\arraystretch}{1.15}
\resizebox{\columnwidth}{!}{
\begin{tabular}{cccccc}
\toprule
 & MQA & G8K & CQA & TQA & SQA \\
\midrule
\multirow{2}{*}{w/o reinit}     & 48.70 & 45.18 & 87.82 & 83.72 & 76.74 \\ [-5.0pt]
                              & (92.86) & (115.91) & (32.95) & (38.27) & (33.89) \\
\midrule
\multirow{2}{*}{w/ reinit}     & 56.95 & 54.98 & 89.56 & 83.45 & 78.19 \\ [-5.0pt]
                              & (91.04) & (100.38) & (33.55) & (39.17) & (35.12) \\
\bottomrule
\end{tabular}
}
\caption{Performance comparison before and after the reinitializing optimizer in every training epoch.
Each row presents accuracy in the first line, with average generated tokens shown in parentheses in the second line.
}
\label{tab:optimizer}
\end{table}

\section{Reinitializing the Optimizer}
\label{appendix:optimizer}

We reinitialized the optimizer after each training epoch to stabilize training, inspired by \citet{deng2024explicitcotimplicitcot}. 
Our implementation uses the AdamW optimizer \cite{loshchilov2018decoupled}, where the first and second moments are gradually updated based on current gradients. 
Consequently, when VARR reduces rationales for certain data points between epochs, the training process could become unstable.
As shown in table \ref{tab:optimizer}, the average 19.32\% performance improvement achieved through optimizer reinitialization extends the findings of \citet{deng2024explicitcotimplicitcot} beyond simple tasks (e.g., multiplication; synthesized dataset) to demonstrate effectiveness across diverse datasets with complex semantic and syntactic reasoning structures.

\section{Zero Warm-up Stage}
\label{appendix:zero-warm}

\begin{table}[h!]
\centering
\renewcommand{\arraystretch}{1.15}
\resizebox{\columnwidth}{!}{
\begin{tabular}{cccccc}
\toprule
warm-up & 0.0 & 0.1 & 0.2 & 0.3 & 0.4 \\
\midrule
\multirow{2}{*}{TQA} & 82.01 & 83.72 & 83.45 & 81.19 & 81.97  \\ [-5.0pt]
                        & (40.11) & (38.27) & (47.10) & (46.55) & (47.97) \\
\midrule
\multirow{2}{*}{CQA} & 90.34 & 89.56 & 87.82 & 87.82 & 86.08 \\ [-5.0pt]
                        & (32.81) & (35.07) & (39.38) & (40.98) & (43.61) \\
\bottomrule
\end{tabular}
}
\caption{Accuracy and average generated tokens across the various warm-up stages.}
\label{tab:optimizer}
\end{table}

To validate whether the warm-up stage sensitivity could limit applicability, we conducted additional experiments extending warm-up stages to zero.
As shown in Table, the zero warm-up stage shows comparable effectiveness and efficiency to the 0.1 warm-up stage, demonstrating the robustness of our method across different application scenarios.
These findings suggest a zero-warm stage can be a good option for various application scenarios, but starting with a 0.1 warm-up stage can boost its stability.

\section{Removal Ratio Analysis on additional datasets.}
\label{appendix:removal-ratio}

\begin{figure}[h!]
    \centering
    \includegraphics[width=1.0\linewidth]{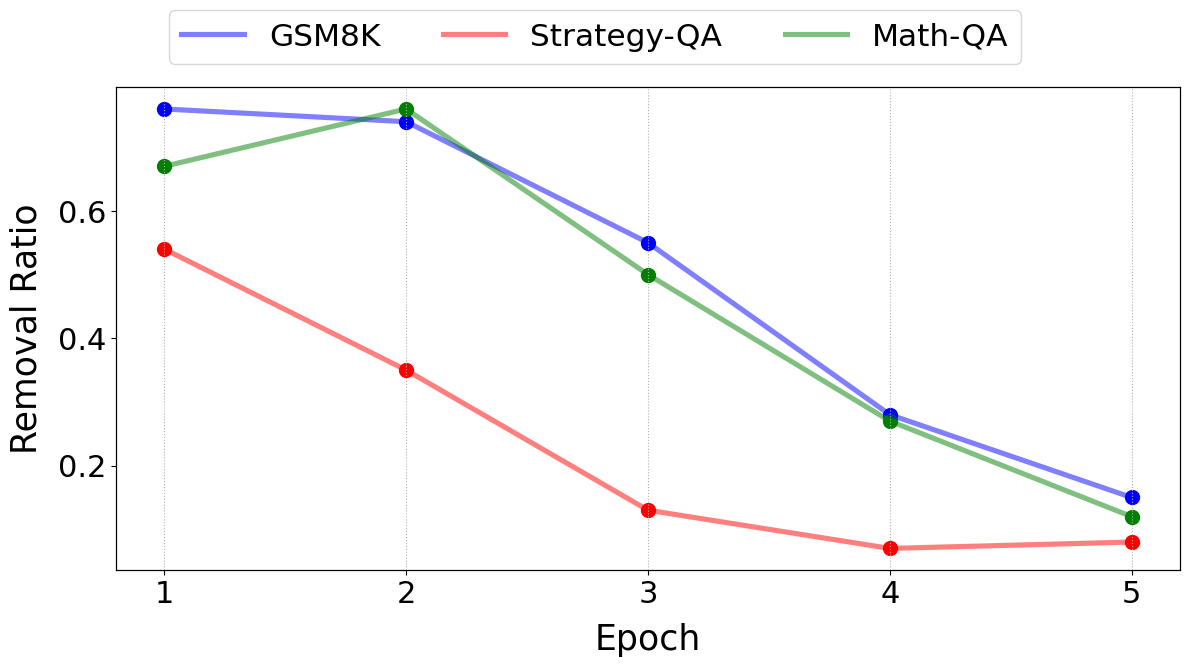 }
    \caption{
    Additional analysis of the Removal ratio on other datasets.
   The Removal ratio, indicated on the y-axis, is calculated by dividing the number of sentences actually removed by the maximum potential removal ratio (i.e., ${size(\{r_i\})/r(t)})$.
    The x-axis represents the index of the epochs.}
    \label{fig:app-removal}
\end{figure}

In the GSM8K, StrategyQA, and MathQA datasets, similar trends are observed as in Section \ref{ab-removal-analysis}. 
Additionally, we observe that our method reduces fewer sentences in later training stages, demonstrating that \textit{verbosity} preserves essential reasoning sentences even when larger $r(t)$ encourages substantial reduction.
Notably, in StrategyQA, the Removal ratio converges more rapidly than in other datasets. 
This is attributed to the dataset's characteristics, where the sentences within the rationale predominantly list simple information rather than forming complex, interrelated structures. 
Consequently, redundant sentences are rapidly removed, leading to a rapid decrease in further removal activities.

\section{Exploration on Complex Task}
\label{appendix:complex}

\begin{table}[h!]
\centering
\small
\begin{tabular}{ccc}
\toprule
 & Exp-CoT & VARR+ \\
\midrule
Acc	& 9.01 & 10.27  \\ 
\midrule
Gen Tok & 300.17 & 252.00 \\
\bottomrule
\end{tabular}
\caption{Performance comparison between Exp-CoT and VARR+ on MATH datasets.}
\label{tab:complex-task}
\end{table}

To further validate that our method remains effective on complex tasks, we conducted additional experiments with the MATH \citep{hendrycksmath2021} dataset. 
Considering computational constraints, we selected 500 samples from the math subset of OpenMathInstruct-2 \citep{toshniwal2024openmath2} for the training set and evaluated on the official test set.
As described in Table \ref{tab:complex-task}, our results on MATH further demonstrate the versatility and future applicability of VARR+ on more complex tasks.

\section{Training Time Analysis}
\label{appendix:training-time}

\begin{table}[h!]
\centering
\small
\begin{tabular}{ccc}
\toprule
Dataset & GSM8K (7K) & TQA (2K) \\
\midrule
Training Time ($\uparrow$) &	0.18 & 0.30  \\ 
\midrule
Token Efficiency ($\downarrow$)& 0.81 & 0.61 \\
\midrule
Accuracy ($\uparrow$)  & 0.98 & 1.01  \\ 
\bottomrule
\end{tabular}
\caption{Training time analysis on GSM8K and TQA. All scores are calculated as the value for Explicit-CoT divided by the value for VARR+ (i.e., Explicit-CoT/VARR+).}
\label{tab:complex-task}
\end{table}

As VARR computes the likelihood of each sentence to ensure robust performance, training time inevitably increases as dataset size grows. 
We therefore quantitatively measure its efficiency on GSM8K and TQA relative to Explicit-CoT. 
Although there is a trade-off in training time, VARR+ improves token efficiency while preserving accuracy. 
Since training occurs only once while the improved efficiency benefits persist continuously afterward in test-time, we believe the additional training cost is justified.

\section{Qualitative Analysis}
\label{appendix:qualitative}
Table \ref{tab:qualitative_example} illustrates the generation outputs for each method on the GSM8K test set.
While Explicit-CoT generates a complete reasoning path, its redundant generation of `needs a 3 piece place setting' introduces hallucinations into the reasoning process, ultimately leading to an incorrect answer. 
ICoT-SI's token-level elimination of reasoning paths, on the other hand, results in generations that lack sufficient reasoning capacity.
Additionally, Coconut's predefined hidden-states fail to compress the complete reasoning path in its token-level compression approach.
In contrast, VARR+ demonstrates efficient generation by concisely producing `\$15.00' in a single sentence, achieving both effectiveness and efficiency in its output.
Additional qualitative examples on the MathQA dataset are outlined in Table \ref{tab:qualitative_example-mqa}.

\begin{table*}[ht!]
    \centering
    \begin{tabular}{|p{\textwidth}|}
        \hline
        \rowcolor{gray!20} \textbf{Question} \\
        \hline
        Avery needs to buy a 3 piece place setting (dinner \& salad plate and a bowl) for her holiday dinner. She's having 12 people over for dinner. If the dinner plates cost \$6.00 each and bowls each cost \$5.00 and the salad plates cost \$4.00, how much will she spend on place settings? \\
        \hline
        
        \rowcolor{gray!20} \textbf{Explicit-CoT} \\
        \hline
        \textcolor{red!50}{She's having 12 people over and needs a 3 piece place setting for each so she needs 12*3 = 36 place settings.} \\ The dinner plates cost \$6.00 each and she needs 36 of them so that comes to 6*36 = \$216.00. \\ The bowl costs \$5.00 each and she needs 36 of them so that comes to 5*36 = \$180.00. \\ The salad plates cost \$4.00 each and she needs 36 of them so that comes to 4*36 = \$144.00. \\ All total, the place settings will cost \$216 for dinner plates, \$180 for bowls and \$144 for salad plates for a total of 216+180+144 = \$540.00 \\ \#\#\#\# 540 \\
        \hline
        
        \rowcolor{gray!20} \textbf{ICoT-SI} \\
        \hline
        each so 12*6 = \$72.00 She's buying 3 pieces of place settings (dinner plates, salad plates and bowls) for \$72.00 each so 3*72 = \$216.00 
        \\ \#\#\#\# 216 \\
        \hline

        \rowcolor{gray!20} \textbf{Coconut} \\
        \hline
        <|start-latent|><|latent|><|latent|><|latent|><|end-latent|>
        \\ \#\#\#\# 4.00 \\
        \hline
        
        \rowcolor{gray!20} \textbf{VARR+} \\
        \hline
        \textcolor{red!80}{She needs 12 place settings and each setting costs \$15.00} so she will spend 12*15 = \$180.00 \\ \#\#\#\# 180 \\
        \hline
        
        \rowcolor{gray!20} \textbf{Ground Truth Answer} \\
        \hline
        \#\#\#\# 180 \\
        \hline
    \end{tabular}
    \caption{Qualitative Example: Generation outputs on GSM8K test set.}
    \label{tab:qualitative_example}
\end{table*}

\begin{table*}[ht!]
    \centering
    \begin{tabular}{|p{\textwidth}|}
        \hline
        \rowcolor{gray!20} \textbf{Question} \\
        \hline
        The cost price of 20 articles is the same as the selling price of $x$ articles. if the profit is 25\%, find out the value of x. \\ a) 13, b) 14, c) 15, d) 16, e) 17 \\
        \hline
        
        \rowcolor{gray!20} \textbf{Explicit-CoT} \\
        \hline
        explanation : let the cost price 1 article = rs . 1 \\
        cost price of $x$ articles = rs . $x$ \\ 
        selling price of x articles = 125\% of $x$ = rs . ( 125 / 100 $x$ ) = rs . ( 25 / 20 $x$ ) = rs . 5 / 4 $x$ \\
        = > $x$ = 20 \\
        \textcolor{red!50}{cost price of 20 articles = rs . 20 selling price of 20 articles = rs . ( 25 / 4 $\times$ 20 ) = rs . ( 25 / 4 $\times$ 20 ) / 20 = rs . 1} \\
        = > $x$ = 20 answer : option a 
        \\ \#\#\#\# a \\  
        \hline
        
        \rowcolor{gray!20} \textbf{ICoT-SI} \\
        \hline
        of $x$ articles = $x$ \\
        s.p of $x$ articles = rs . 20 \\
        gain = rs . ( 20 - $x$ ) \\ 
        gain \% = 25 \% of $x$ = rs . ( 20 - $x$ ) / $x$ $\times$ 100 = 25 \\
        = > 20 - $x$ = 5 $x$ = > 4 $x$ = 20 - $x$ = > 5 $x$ = 20 - $x$ = > 20 = 5 $x$ = > 200 = 5 $x ^ 2$ = > 400 = 5 $x ^ 2$ = > 400 / 5 = $x ^ 2$ = > $x ^ 2$ = 80 = > x = 10 or 10 . c \\
        \#\#\#\# c \\
        \hline

        \rowcolor{gray!20} \textbf{Coconut} \\
        \hline
        <|start-latent|><|latent|><|latent|><|latent|><|end-latent|>
        \\ \#\#\#\# c \\
        \hline
        
        \rowcolor{gray!20} \textbf{VARR+} \\
        \hline
        \textcolor{red!80}{explanation : let the cost price of $x$ articles = $x$} \\ 
        \textcolor{red!80}{s.p of $x$ articles = 20} \\
        20 = 1.25 $x$ = > $x$ = 16 option d. 
        \\ \#\#\#\# d \\
        \hline
        
        \rowcolor{gray!20} \textbf{Ground Truth Answer} \\
        \hline
        \#\#\#\# d \\
        \hline
    \end{tabular}
    \caption{Qualitative Example: Generation outputs on MathQA test set.}
    \label{tab:qualitative_example-mqa}
\end{table*}

\end{document}